\documentclass[10pt,twocolumn,letterpaper]{article}

\pdfoutput=1

\usepackage[applications]{wacv}      

\usepackage{graphicx}
\usepackage{float}
\usepackage{amsmath}
\usepackage{amssymb}
\usepackage{booktabs}
\usepackage{multirow}

\usepackage[pagebackref,breaklinks,colorlinks]{hyperref}

\usepackage[capitalize]{cleveref}
\crefname{section}{Sec.}{Secs.}
\Crefname{section}{Section}{Sections}
\Crefname{table}{Table}{Tables}
\crefname{table}{Tab.}{Tabs.}

\def\wacvPaperID{6} 
\def\confName{WACI}
\def\confYear{2025}

\begin{document}

\title{Fashionability-Enhancing Outfit Image Editing\\with Conditional Diffusion Models}

\author{Qice Qin\\
Waseda University\\
Tokyo, Japan\\
{\tt\small qinqice@toki.waseda.jp}
\and
Yuki Hirakawa\\
ZOZO Research\\
Tokyo, Japan\\
{\tt\small yuki.hirakawa@zozo.com}
\and
Ryotaro Shimizu\\
ZOZO Research\\
Tokyo, Japan\\
{\tt\small ryotaro.shimizu@zozo.com}
\and
Takuya Furusawa\\
ZOZO Research\\
Tokyo, Japan\\
{\tt\small takuya.furusawa@zozo.com}
\and
Edgar Simo-Serra\\
Waseda University\\
Tokyo, Japan\\
{\tt\small ess@waseda.jp}
}

\maketitle

\begin{abstract}
Image generation in the fashion domain has predominantly focused on preserving body characteristics or following input prompts, but little attention has been paid to improving the inherent fashionability of the output images. This paper presents a novel diffusion model-based approach that generates fashion images with improved fashionability while maintaining control over key attributes. Key components of our method include: 1) fashionability enhancement, which ensures that the generated images are more fashionable than the input; 2) preservation of body characteristics, encouraging the generated images to maintain the original shape and proportions of the input; and 3) automatic fashion optimization, which does not rely on manual input or external prompts. We also employ two methods to collect training data for guidance while generating and evaluating the images. In particular, we rate outfit images using fashionability scores annotated by multiple fashion experts through OpenSkill-based and five critical aspect-based pairwise comparisons. These methods provide complementary perspectives for assessing and improving the fashionability of the generated images. The experimental results show that our approach outperforms the baseline Fashion++ in generating images with superior fashionability, demonstrating its effectiveness in producing more stylish and appealing fashion images.
\end{abstract}


\section{Introduction}
\label{sec:intro}

\begin{figure}[t]
   \centering
   \includegraphics[width=\linewidth]{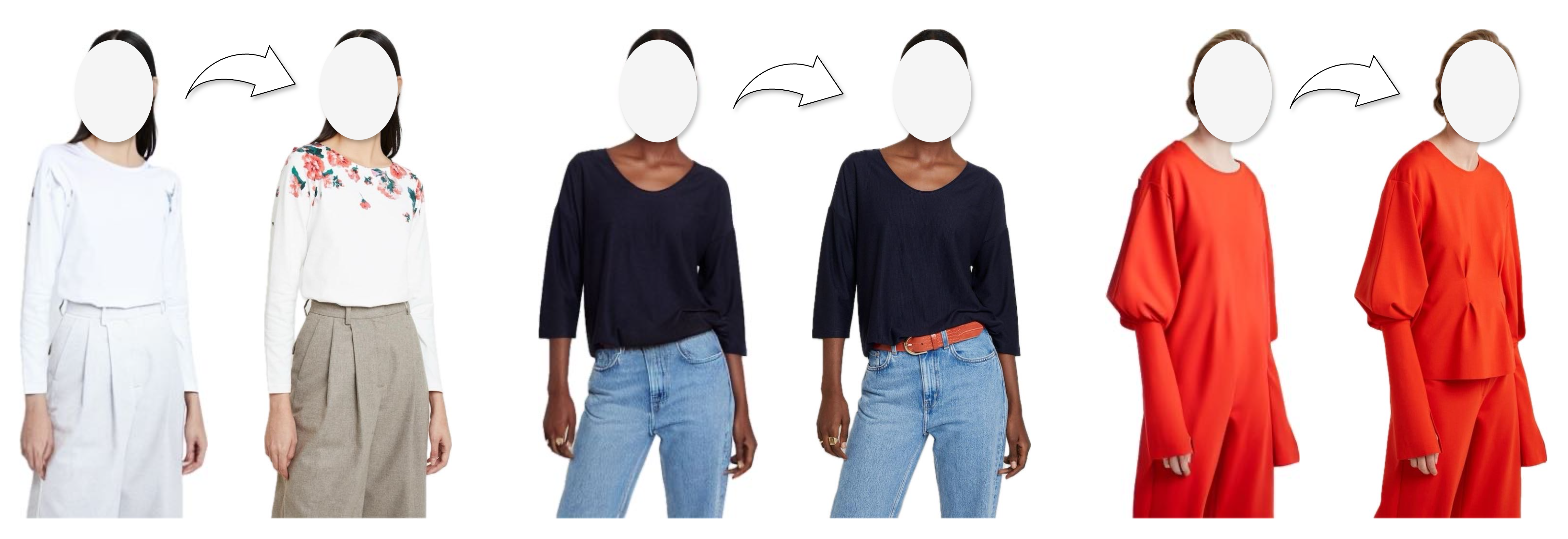}
   \caption{Illustrations of minimal edits to enhance fashionability. 1) Modifying white pants to brown and adding floral patterns to the blouse (left). 2) Adding a belt to the outfit as an accessory (center). 3) Adjusting the shape and size of a loose-fitting red jumpsuit to a more form-fitting and tailored version (right).}
  \label{fig:minimaledit}
\end{figure}

\begin{figure*}[t]
\centering
\includegraphics[width=\textwidth]{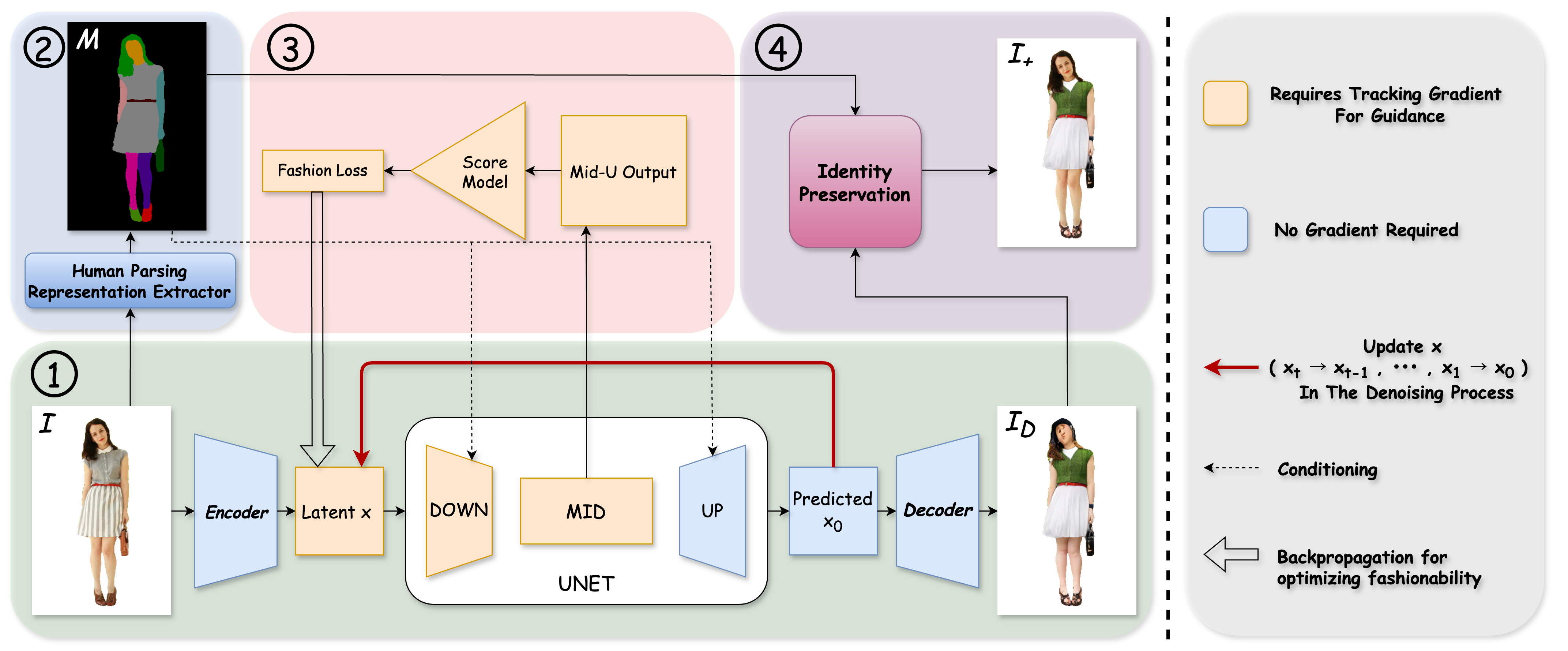}
  \caption{Overview of our proposed approach for conditional fashion image generation. Our approach consists of four main components: 1) a diffusion-based generation module that applies a diffusion process to iteratively refine the input image into a fashion-enhanced output while maintaining visual coherence; 2) a human parsing representation extractor that generates segmentation maps from the input image; 3) a mid-U classifier that processes mid-level UNet outputs to compute the fashion loss, which is fed back to the latent representation for fashionability optimization; and 4) an identity preservation process that combines the generated output with the segmentation map to restore the original subject's head onto the new image.}
  \label{fig:overview}
\end{figure*}

The fashion industry is continuously evolving, with remarkable advancements in AI technology~\cite{fashion0,fashion1,fashion2,fashion3,fashion4,fashion5,fashion6,fashion7,fashion8,fashion9,fashion10,fashion11}. One of the challenges in fashion e-commerce is the lack of expert advice that consumers would typically receive in physical stores, making it essential to develop features that support users' decision-making. In particular, recent developments in virtual try-on systems~\cite{viton0,viton1,viton2,viton3,viton4,viton5,viton6,viton7,viton8,viton9,viton10,viton11,viton12} and clothed human generation models~\cite{human1,huamn2,huamn3,huamn4,huamn5}, enable users to preview clothing on themselves without visiting physical stores. These models mainly focus on preserving body characteristics or following predefined prompts to generate fashion images~\cite{baldrati2023multimodal,lampe2024dicti}. However, they often overlook enhancing the inherent fashionability of the generated images, which limits their effectiveness in supporting users unfamiliar with fashion.

To address this issue, we introduce an approach to enhance the fashionability of input images, helping users and designers make better fashion decisions. Figure~\ref{fig:minimaledit} shows the concept of enhancing fashionability through minimal edits.\footnote{The base images (right) are sourced from the VITON-HD dataset~\cite{viton12}, and the negative examples are made by InstructPix2Pix~\cite{instructpix2pix}.} It showcases three pairs of images that demonstrate how minor adjustments, such as modifying colors and patterns, adding accessories and changing clothing styles, can improve overall fashionability.

In this paper, we propose a novel approach to enhance the fashionability of generated images by leveraging intermediate latent features and structured conditioning during the denoising process of a latent diffusion model~\cite{Rombach_2022_CVPR}. Our method incorporates a two-fold strategy: (1) utilizing an intermediate feature-based classifier to guide the diffusion process and (2) integrating segmentation map-based conditioning to ensure structural fidelity in the generated images.

First, We train a classifier on an expert-annotated dataset to predict fashionability scores and incorporate it into the generation process to produce images with enhanced fashionability.
By introducing a guidance model trained on expert-annotated fashionability data, our approach enables more reliable generation compared to methods like Fashion++, which relies on artificially generated negative examples.

Additionally, we train a diffusion model conditioned on segmentation maps using images from WEAR~\cite{app_wear}\footnote{The largest fashion outfit-sharing application in Japan.} to preserve the original clothing and body shape characteristics.
Unlike previous methods~\cite{baldrati2023multimodal,lampe2024dicti} that use detailed prompts describing garment styles and colors, our approach does not require specific prompts during training and inference.
This allows our model to generate more fashionable images without external user inputs.
Finally, we use the diffusion pipeline augmented with Classifier Guidance.
We calculate the fashionability loss of the latent features at each step using the trained classifier. This fashionability loss guides the diffusion model toward generating images with higher fashionability.

To assess the effectiveness of our method, we incorporate two additional evaluation models. The first is an OpenSkill~\cite{Joshy2024}-based model, which leverages annotated pairwise comparisons to assign continuous fashionability scores, providing a quantitative evaluation framework. The second is a classifier that evaluates fashionability across five critical dimensions—\textit{cleanliness}, \textit{harmony}, \textit{shilhouette}, \textit{styling}, and \textit{trendiness}—offering a more comprehensive perspective on fashion perception.

Extensive experiments demonstrate that our method outperforms the baseline Fashion++~\cite{hsiao2019fashion++} in generating fashionable images.
By combining a reliable, expert-annotated dataset with recent advancements in machine learning techniques, our research provides a robust framework for automatically enhancing the fashionability of generated images.

The main contributions of this study are summarized as follows: 
\begin{itemize}
    \item Our approach combines ControlNet with Classifier Guidance, allowing the model to generate highly fashionable images while preserving the input image’s body and clothing shape characteristics.

    \item We develop two complementary fashionability prediction models to assess the fashionability of fashion images. The first model is trained on a dataset with absolute fashion scores annotated by fashion experts. The second is trained on expert evaluations across five critical aspects of fashionability, providing a comprehensive framework for evaluation.
    
    \item Our method undergoes evaluation, including quantitative metrics and visual quality assessment, demonstrating its superiority in generating highly fashionable images compared to the existing method.
\end{itemize}


\section{Related Work}
\label{sec:formatting}

\subsection{Fashion image generation}

Fashion image generation has been a prominent area of research, and various methods have been proposed to create realistic and stylish fashion images.
Hsiao et al.~\cite{hsiao2019fashion++} introduced Fashion++, a data-driven approach that uses a BicycleGAN~\cite{zhu2017toward} framework to enhance the fashionability of existing outfits with minimal edits. Specifically, Fashion++ makes subtle but effective changes, such as modifying colors, adjusting textures, or removing unnecessary accessories, to create visually appealing outfits. The model is trained on a dataset of outfit images, using a loss function that combines reconstruction loss and feature matching loss to ensure the edits are both realistic and fashion-forward. As the sole baseline in this field, Fashion++ demonstrates the potential of leveraging generative models to improve clothing aesthetics in a data-driven manner.
However, it has notable limitations, such as generating low-resolution images and relying on augmented negative samples (i.e., slightly altered versions of the original images) for training. This reliance on artificially generated data may overlook the subjective and multifaceted nature of fashionability, failing to fully capture its nuanced aspects. In contrast, our approach leverages real annotations in order to modify images with increased fashionability with higher fidelity.

On the other hand, Lampe et al.~\cite{lampe2024dicti} proposed DiCTI, a diffusion-based clothing designer guided by textual input, focusing on generating fashion images based on user-provided descriptions.
This method showcases the potential of diffusion models in creating diverse and stylish clothing designs but relies on detailed textual prompts. On the contrary, our approach can utilize diffusion models to improve fashionability without needing additional text prompts.

\subsection{Fashionability Prediction}

Fashionability prediction has been challenging due to its subjective nature and the numerous factors influencing fashion perception.
Prior research has developed datasets by engaging human annotators in comparing pairs of images featuring individuals dressed in diverse outfits and selecting the one deemed more aesthetically pleasing ~\cite{8575482,Neuberger2018}. Subsequently, these datasets were utilized for training and assessing deep learning models capable of evaluating fashion aesthetics.

Simo-Serra et al.~\cite{simo2015neuroaesthetics} made significant contributions in this area by developing a model for analyzing the perception of fashionability. 
Their model considers various factors, including outfit styles, individual user characteristics, the background of the photograph, and the resulting fashionability score. 
It provides detailed suggestions to users, such as recommending specific garment adjustments or background changes to enhance their perceived fashionability.

Hirakawa et al.~\cite{niau_hirakawa} conducted an in-depth analysis of the correlation between human judgments and predictions made by GPT-4~\cite{openai2024gpt4}. They demonstrated that while GPT-4 shows some correlation with human perception, GPT-4 still struggles to achieve human-level performance. This highlights the inherent difficulty in predicting fashionability, even for GPT-4 in zero-shot. 

Motivated by these findings, we adopt an alternative approach for fashionability prediction that emphasizes precision and human alignment. We prepare small but carefully annotated datasets and train evaluation models on them, aiming to better capture the nuanced and subjective aspects of fashionability. Our approach avoids the pitfalls of the general-purpose model and enables us to evaluate the fashionability of images in a more reliable and comprehensive manner.


\section{Methodology}

\begin{figure*}
   \centering
   \includegraphics[width=\textwidth]{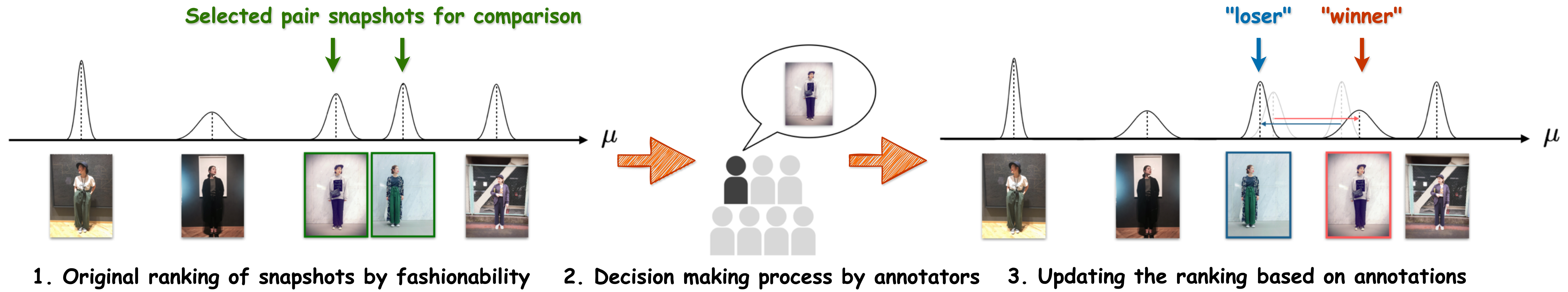}
   \caption{Illustration of the OpenSkill-based fashionability scoring process. The first part shows images ranked along an axis, representing their average fashionability scores with associated normal distributions. The second part depicts the annotation process, where human evaluators provide pairwise comparisons to determine which image is more fashionable. The third part reflects the updated scores and distributions of the images, adjusted based on the annotations.}
  \label{fig:openskill}
\end{figure*}

Figure~\ref{fig:overview} illustrates the four main modules of our proposed method for conditional fashion image generation: \textbf{1) Diffusion-based generation module}: The input image is processed through a diffusion model to produce a new output image to enhance its fashionability. \textbf{2) Human parsing representation extractor}: Segmentation maps are generated from the input image to preserve the subject's structural and proportional details. \textbf{3) Fashionability feedback loop}: The mid-U output from the UNet is evaluated by a mid-U classifier, which calculates a fashion loss. This loss is then used to refine the latent representation in the diffusion process, guiding the model towards generating images with higher fashionability. \textbf{4) Identity preservation}: To ensure the output retains the subject's identity, the generated image and the segmentation map are combined to restore the original subject's head onto the new image.

Our system's modular design allows it to simultaneously enhance fashionability, maintain body characteristics, and preserve the subject's identity, making it both practical and interpretable.

\subsection{Dataset and Preprocessing}

Acquiring annotations for fashionability is inherently challenging due to its highly subjective and nuanced nature. Our study utilizes two datasets: the OpenSkill-based and 5-scores-based datasets, each tailored for specific approaches discussed in the following subsections. Each image in our datasets was carefully evaluated by annotators across multiple aspects, making the annotation process both labor-intensive and costly. As a result, building a large-scale dataset is impractical within this context. Despite the relatively small size of our datasets, their detailed and high-quality annotations ensure their utility in advancing the research on fashion image generation and fashionability evaluation. Figure~\ref{fig:dataset} illustrates the construction and division of these datasets into training and testing subsets. This visualization provides a comprehensive overview of the datasets used throughout the study.

\subsubsection{OpenSkill-based Annotations}
\label{subsec:openskill}

We adopt OpenSkill~\cite{Joshy2024}, a framework originally designed for estimating player skills in competitive games. OpenSkill models each player’s skill as a normal distribution characterized by a mean ($\mu$) and standard deviation ($\sigma$), which are updated based on the outcomes of pairwise matchups. In our case, we treat fashion snapshots as players and their aesthetic scores as skill ratings.

The annotation process consists of three main steps, as illustrated in Figure~\ref{fig:openskill}. Initially, the fashion snapshots are represented as normal distributions with prior scores. First, we conduct a matchup by sampling two snapshots based on the expected information gain from their comparison. Second, an annotator is randomly chosen to evaluate the pair and determine the more fashionable outfit. Third, the OpenSkill algorithm updates the parameters of both snapshots based on the annotator's decision. These steps are iterated until the system converges to stable scores for all snapshots. Following a similar methodology to Kiapour et al.\cite{kiapour2014hipster}, which utilized TrueSkill\cite{herbrich2006trueskill}, a predecessor of OpenSkill, we constructed image pairs for comparison by identifying the image with the lowest match rate and pairing it with a counterpart predicted to have a high likelihood of resulting in a draw, as determined by the OpenSkill scores. To ensure the accuracy of the annotations, we prepared two independent annotator groups and repeated the above procedure until Spearman’s rank correlation coefficient between the groups reached saturation. The final coefficient is 0.814, indicating that our dataset is highly reliable.\footnote{The details of our method for creating datasets are shown in \cite{niau_hirakawa}.}

Our dataset comprises 6,000 fashion snapshots from WEAR~\cite{app_wear}, annotated by hundreds of human evaluators. These snapshots underwent 219,488 pairwise comparisons, forming the foundation of our scoring process. Out of the 6,000 images, 5,000 were allocated to train the mid-U classifier to guide the generation process, and the remaining 1,000 images were reserved to train an OpenSkill-based model for fashionability prediction as well as to test the mid-U classifier. Additionally, 500 images from the original 5,000 training set were set aside to test the OpenSkill-based model.

\subsubsection{5-Dimension Annotations}

Additionally, we utilized a second proprietary dataset of 831 images, each evaluated by annotators across five aspects: \textit{cleanliness}, \textit{harmony}, \textit{shilhouette}, \textit{styling}, and \textit{trendiness}, which were determined as essential by professionals with extensive fashion knowledge. The five dimensions and their evaluation criteria are as follows: \textbf{1) Cleanliness}: evaluates whether the outfit appears clean and well-maintained. \textbf{2) Harmony}: assesses whether the wearer's hairstyle, makeup, and skin tone match the overall outfit. \textbf{3) Silhouette}: examines the balance of proportions and fit relative to the wearer’s body. \textbf{4) Styling}: measures the sense of styling in how the outfit is worn, including combining colors and patterns. \textbf{5) Trendiness}: determines whether the outfit reflects current fashion trends. 

We collected annotations from experts using pairwise comparisons, in which annotators were presented with pairs of images and asked to determine which image performed better in a specific aspect. This process resulted in a total of 180,548 pairwise comparison transactions, distributed as follows: Cleanliness (32,628), Harmony (35,265), Silhouette (34,833), Styling (34,461) and Trendiness (43,361). These transactions provide a robust and detailed foundation for modeling the nuanced evaluation of fashion elements across these dimensions.

We normalized each score so that it takes an integer value ranging from one to five and enables consistent comparisons across all aspects. Furthermore, we calculated the average score across the five aspects to derive an overall measure of fashionability. We rounded it to the nearest integer, thus obtaining a single, unified fashionability score for each image. The overall metric and individual ones allow us to score both the general impression and specific strengths or weaknesses of the fashionability of outfits.

This dataset was divided to train our 5-scores-based model that assesses fashionability along these dimensions, with 89.9\% of images used for training and the remaining 10.1\% of images used for testing. During the training process, the dataset was initially divided into training, validation, and testing subsets, with 80\% used for training, 10\% for validation, and 10\% for testing. However, to fully utilize the limited dataset, the validation subset was merged into the training set for the final model training for 30 epochs.

\begin{figure}
   \centering
   \includegraphics[width=\linewidth]{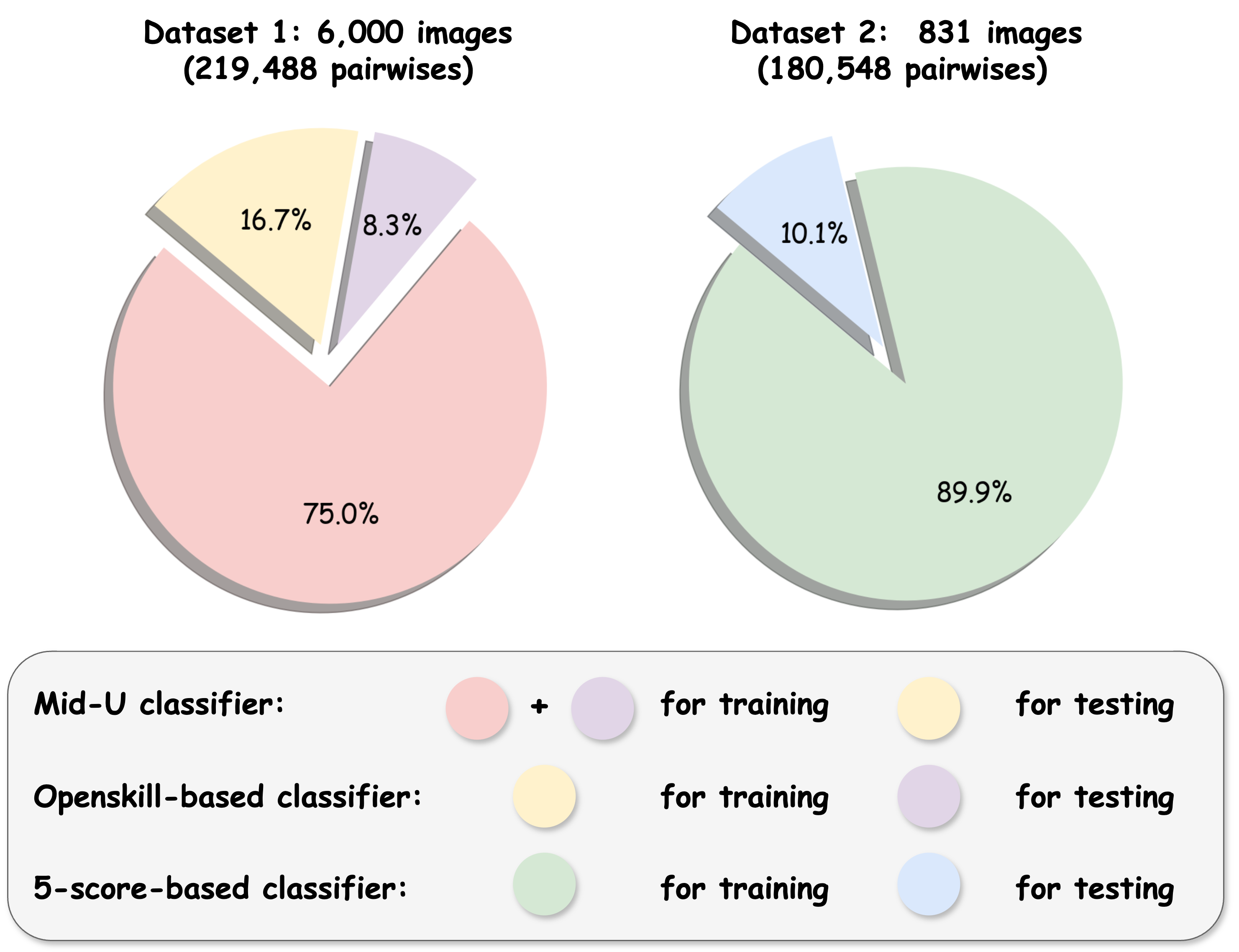}
   \caption{Construction of the dataset. It shows how we divided them for each model's training and testing.}

  \label{fig:dataset}
\end{figure}

\subsection{Base Architecture}

Our approach leverages ControlNet as the backbone model to incorporate additional control into the image generation process, ensuring the generated images adhere to specific desired attributes.
We use segmentation maps as the primary conditioning input for ControlNet.
These segmentation maps are generated from fashion images by an out-of-box human parsing representation extractor~\cite{li2020self}, which helps capture the overall structure of the original images, such as layout and pose.
Specifically, ControlNet is conditioned on segmentation maps \(\mathbf{M}\) of the input image \(\mathbf{I}_{\text{img}}\).
During training, the forward process incrementally adds noise to the input image \(\mathbf{I}_{\text{img}}\), resulting in a noisy image \(\mathbf{x}_t\), where \(t\) indicates the time step, the number of times noise is added.
The network \(\epsilon_\theta\) is trained to predict the noise added to the noisy image \(\mathbf{x}_t\) using the time step \(t\) and the segmentation maps \(\mathbf{M}\) as conditions.
The objective function for training ControlNet is defined as follows:
\begin{equation}
\mathcal{L}_{\text{control}} = \mathbb{E}_{\mathbf{I}_{\text{img}}, t, \mathbf{M}, \epsilon \sim \mathcal{N}(\mathbf{0}, \mathbf{\Sigma_{I}})} \left[ \left\| \epsilon - \epsilon_\theta \left(\mathbf{x}_t, t, \mathbf{M}\right) \right\|_2^2 \right] \;,
\end{equation}
\noindent where \(\mathcal{L}_{\text{control}}\) represents the overall training objective for the diffusion model.
This objective guides the fine-tuning process of diffusion models with ControlNet to ensure that the generated images retain the original characteristics of clothing and body shape. In our implementation, we fine-tuned all parameters of ControlNet, leveraging the full model capacity to adapt it to our dataset.

\subsection{Classifier Guidance}

Integrating Classifier Guidance into the diffusion model is a crucial aspect of our approach. The Classifier Guidance helps steer the diffusion process toward generating images that are more fashionable by utilizing a pre-trained classifier that provides feedback at each step of the image generation. Specifically, we use Mid-U Guidance~\cite{MidU} rather than the original Classifier Guidance technique~\cite{dhariwal2021diffusion}. The key difference between Mid-U Guidance and the original Classifier Guidance is that, while the original technique operates directly in pixel space and requires gradients to be backpropagated through both the classification model and the decoder of a Variational Autoencoder~\cite{kingma2013auto} (for latent diffusion models), Mid-U Guidance leverages intermediate features from the UNet's encoder. This enables more efficient guidance by significantly reducing both computation and memory usage. We chose Mid-U Guidance because it is more computationally efficient, allowing for faster processing while still providing high-quality guidance in the fashionability enhancement process.

During the diffusion process, the UNet model produces intermediate latent features, specifically from the mid-block of the network, denoted as \( \epsilon_{\text{mid}}(\mathbf{x}_{t,i}) \). These features are extracted and used as input for a pre-trained fashion classifier \( f_\phi \). The classifier categorizes these intermediate features into three distinct fashionability classes: high, medium, and low. The dataset~\ref{subsec:openskill} used for this classification consists of 6,000 images annotated with continuous scores. We normalized each score to an integer value ranging from one to three and it is evenly divided into these three classes, with 2,000 images per class. This setup allows us to explicitly incorporate the fashionability score into the guidance process by leveraging the inherent order of the class labels. Unlike standard classification tasks, where class labels are independent, our approach assigns greater importance to higher fashionability scores to align to enhance fashionability. 

The fashionability loss \( \mathcal{L}_{\text{fashion}} \) is defined as:
\begin{equation}
    \mathcal{L}_{\text{fashion}} = -\frac{1}{N} \sum_{i=1}^{N} \sum_{j=1}^{3} \frac{\exp\left(f_\phi\left(\epsilon_{\text{mid}}\left(\mathbf{x}_{t,i}\right)\right)_j\right)}{\sum_{k=1}^{3} \exp\left(f_\phi\left(\epsilon_{\text{mid}}\left(\mathbf{x}_{t,i}\right)\right)_k\right)} \cdot j, \; 
\end{equation}
\noindent where \( f_\phi \) represents the fashion classifier, which outputs probabilities for three fashionability classes, and \( j \) is the class label (1: low, 2: medium, 3: high). Here, the softmax function is explicitly defined to provide a probability distribution over the fashionability classes. The weighted summation of class probabilities ensures that the loss prioritizes higher fashionability scores, effectively serving as a soft ordinal regression mechanism.

The computed fashionability loss is then back-propagated through the network, where the gradient of the fashionability loss with respect to the latent representations \( \mathbf{x}_{t,i} \) is then computed as:
\begin{equation}
\nabla_{\mathbf{x}_{t,i}} \mathcal{L}_{\text{fashion}} = \lambda \frac{\partial \mathcal{L}_{\text{fashion}}}{\partial \mathbf{x}_{t,i}}  ,
\end{equation}
where \( \lambda \) is a scalar value representing the guidance loss scale. 
Finally, the latents are updated iteratively using the computed gradient and scaled by \(\sigma^2\), where \(\sigma\) is the noise standard deviation at the current step \(i\):
\begin{equation}
\mathbf{x}_{t,i+1} = \mathbf{x}_{t,i} - \sigma^2  \nabla_{\mathbf{x}_{t,i}} \mathcal{L}_{\text{fashion}} \;.
\end{equation}
This iterative update ensures that the generated image aligns with the desired fashionability characteristics, as determined by the aesthetic loss function.


\section{Experimental Results}

\subsection{Implementation Details}

We used a pre-trained Stable Diffusion model v1-5~\cite{Rombach_2022_CVPR} to train ControlNet as a base model.
The dataset used for training consisted of 14,983 images from Wear.jp, along with corresponding segmentation maps that annotate the fashion items present in each image, such as tops, bottoms, and accessories.
The image resolution was set to 512 $\times$ 512, and the learning rate was fixed to $10^{-6}$.
In equations (3) and (4), \( \lambda \) , the scaling factor for the guidance loss, was empirically set to 0.1, while \(\sigma\), the noise standard deviation at each iteration, we used the UniPCMultistepScheduler~\cite{zhao2024unipc} to determine the noise schedule.
To optimize memory usage, we set the batch size and gradient accumulation steps to one and four.
Additionally, the training employed an 8-bit Adam optimizer and memory-efficient attention using xformers~\cite{xformers}.
The model was trained for 93,000 steps on four Tesla V100-SXM2-16GB GPUs.
For the Mid-U Classifier, we took as input features of dimension \(1280 \times 8 \times 8\), which are intermediate outputs from the UNet model.
The fashion classifier \( f_\phi \) mentioned in section 3.3 consists of a sequential network that begins with a $3\times3$ convolutional layer, which takes an input of 1,280 channels and reduces it to 256 channels with padding of one, followed by a ReLU activation. A $2\times2$ max-pooling layer succeeds this, and then a second $3\times3$ convolutional layer that reduces 256 channels to 128 channels, also with padding of one, followed by another ReLU activation. The output is passed through an adaptive average pooling layer, which reduces the spatial dimensions to $2\times2$, and then flattened to a 512-dimensional vector. A fully connected layer then reduces this to 64 dimensions with a ReLU activation. Finally, an output layer maps the 64-dimensional vector to three outputs representing the three fashionability classes. The Mid-U Classifier was trained on the OpenSkill-based dataset with the train-test split shown in Figure~\ref{fig:dataset}, using a cross-entropy loss function. We trained the model for 50 epochs with a batch size of 32 and used the Adam optimizer with its default parameters, setting the learning rate to $10^{-4}$.

\begin{figure*}
  \includegraphics[width=\textwidth]{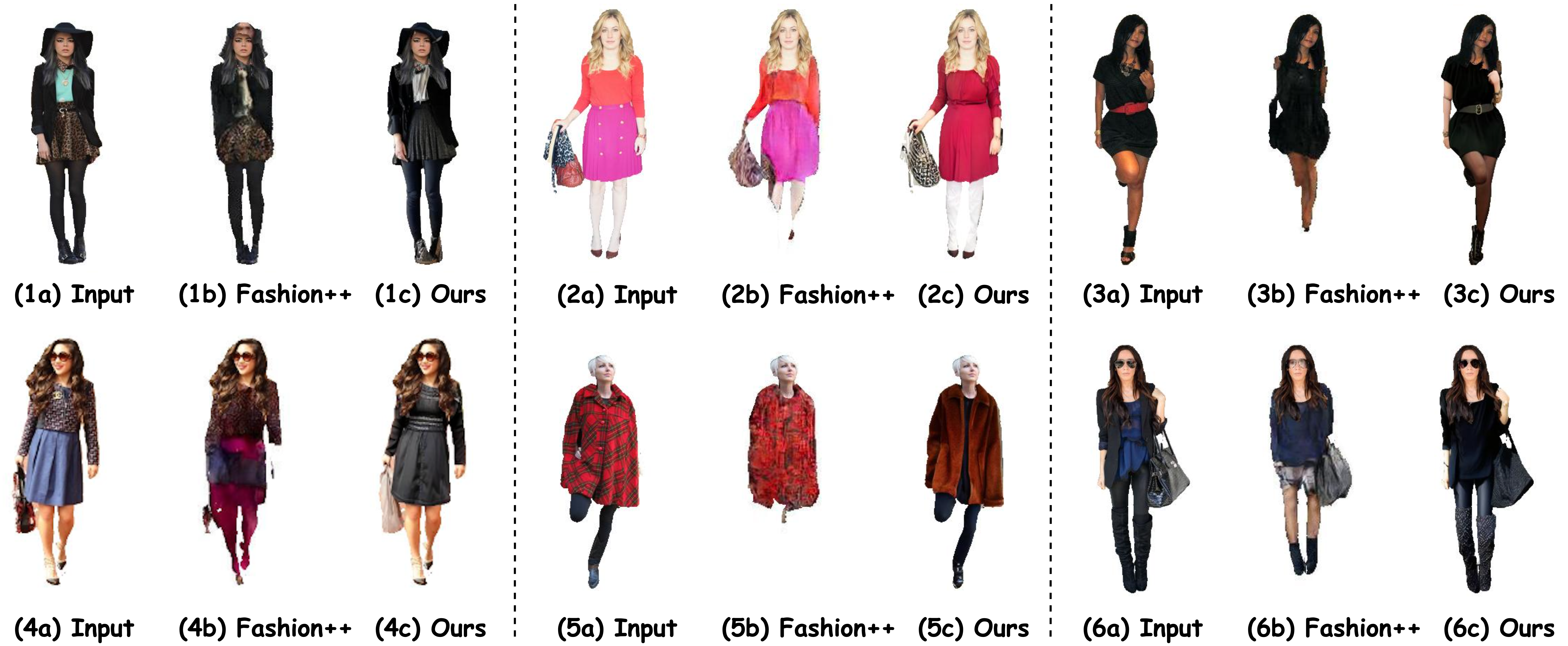}
   \caption{Qualitative comparison of our proposed approach and Fashion++ for the same input.}
  \label{fig:comparison}
\end{figure*}

\subsection{Fashionability Prediction and Evaluation Metrics}

The fashionability prediction model provides quantitative evaluations of generated images to compare different fashion image generation methods.
We developed two distinct models for fashionability prediction: \textbf{1) OpenSkill-based model}, trained on a dataset of 1,000 images categorized into high, medium, and low fashionability with a ResNet-50~\cite{resnet} classifier pre-trained on ImageNet~\cite{krizhevsky2012imagenet}.
We emphasize that those images do not overlap with those used to train the classifier in the generative model.
\textbf{2) 5-scores-based model}, trained on the other proprietary and confidential dataset of 831 images, each evaluated by experts across five aspects: trend, silhouette, sense, cleanliness, and dressing.
A ResNet classifier was trained to predict the average scores as fashionability (very high, high, medium, low, and very low fashionability) across the five categories.
While the OpenSkill-based model offers a broad evaluation based on a large dataset, the 5-scores-based model provides a detailed, expert-driven evaluation, ensuring a solid assessment of fashionability enhancement across diverse criteria.
To evaluate the performance of our method, we employed several key metrics that reflect the quality and fashionability of the generated images:
\textbf{1) Accuracy of Fashionability Prediction Models:} We measured the accuracy of our ResNet classifier trained on two datasets: one with 1,000 images and another with 831 images.
These models provide a baseline for assessing the effectiveness of our model in predicting fashion scores.
\textbf{2) Comparison of Fashionability Scores:} We compared the fashionability scores of images generated by our model and those generated by Fashion++.
This comparison assessed the increase or decrease in fashionability scores assigned by two different prediction models.
\textbf{3) Visual Quality of Generated Images:} We qualitatively assessed the visual quality of images generated by our model versus those generated by Fashion++.
This metric focuses on the generated images' clarity, detail, and overall aesthetic appeal.
\textbf{4) User Study:} We conducted a straightforward user study to ask participants if the generated images by Fashion++ and our method are more fashionable than the original images. This study validates whether actual users perceive the generated images as more fashionable.

\begin{table}[t]

\centering
\resizebox{\linewidth}{!}{ 
\begin{tabular}{ccrr}
\toprule
\textbf{Evaluator} &
\textbf{Method} & \textbf{Increased} \text{\small ($\uparrow$)} & \textbf{Decreased} \text{\small ($\downarrow$)} \\
\midrule
\multirow{2}{*}{OpenSkill-based} & Fashion++ & 22\% & 28\% \\
 & Ours & \textbf{56\%} & \textbf{12\%} \\
\midrule
\multirow{2}{*}{5-scores-based} & Fashion++ & 10\% & 20\% \\
 & Ours & \textbf{26\%} & \textbf{12\%} \\
\bottomrule
\end{tabular}
}
\caption{Comparison of fashionability predictions results by OpenSkill-based model and 5-scores-based model on 50 images generated by Fashion++ and our approach with the same input images.
The numbers indicate the ratio of images rated as having increased or decreased fashionability by each model.}

\label{tab:summary_results}
\end{table}


\subsection{Quantitative Results}
The ResNet classifier trained on 1,000 images achieved an accuracy of 60\% in predicting high, medium, and low fashionability categories on a separate test set from the 5,000 images used to train the Mid-U classifier, outperforming the random chance accuracy of 33\%.
The model trained on 831 images with detailed fashion scores achieved an accuracy of 65\%, which is notably higher than the random chance accuracy of 20\%.
We consider these accuracies reasonable, given the highly subjective nature of fashion and its inherent challenges even for human judgment.
In addition, we conducted an evaluation of fashionability predictions for 50 images generated by two methods: Fashion++ and our proposed method.
The results, as summarized in Table~\ref{tab:summary_results}, indicate a significant improvement in the fashionability of images generated by our method.
These results collectively demonstrate the model's ability to effectively enhance the original images' fashionability, surpassing the baseline method, Fashion++.

\subsection{Qualitative Results}
From Figure~\ref{fig:comparison}, it is evident that for images not seen during training, the results produced by Fashion++ tend to make the subjects appear thinner and often fail to generate the complete body shape accurately, with some instances even missing parts of the legs and feet. Additionally, the fabric textures and overall image quality of the clothing generated by Fashion++ are not very clear. In contrast, our method successfully maintains the original body characteristics while generating clear and detailed images of the clothing. This improvement can be attributed to using a supervised classifier trained on our annotated data, which provides precise guidance during generation, and the high-quality image synthesis capability of Stable Diffusion 1.5. These factors enable our model to make subtle adjustments to enhance fashionability, such as changing the color of a belt, the style of clothing, and the overall outfit coordination. These improvements demonstrate the effectiveness of our model in producing high-quality and fashionable images.

\subsection{User Study}
To evaluate and compare the perceived fashionability of images generated by Fashion++ and our proposed approach. We also conducted a simple user study, where we recruited five participants. Initially, they were presented with 50 sets of images, each comprising an original input image, an output image generated by Fashion++, and an output image generated by our method. We inquired from them about the number of images within these 50 sets where the Fashion++ output exhibited an increase in fashionability compared to the input images, as well as the number of images where our method's output demonstrated an enhanced fashionability compared to the input images. In total, we collected 250 responses, and Table ~\ref{tab:user_study} summarizes the results of this study.

\begin{table}[t]

\centering
\resizebox{\linewidth}{!}{ 
\begin{tabular}{ccrr}
\toprule
\textbf{Participants} &
\textbf{Method} & \textbf{Increased} \text{\small ($\uparrow$)} & \textbf{Decreased} \text{\small ($\downarrow$)} \\
\midrule
\multirow{2}{*}{1} & Fashion++ & 7 & 23 \\
 & Ours & \textbf{25} & \textbf{9} \\
\multirow{2}{*}{2} & Fashion++ & 11 & 30 \\
 & Ours & \textbf{31} & \textbf{12} \\
\multirow{2}{*}{3} & Fashion++ & 3 & 36 \\
 & Ours & \textbf{28} & \textbf{8} \\
\multirow{2}{*}{4} & Fashion++ & 9 & 26 \\
 & Ours & \textbf{20} & \textbf{6} \\
\multirow{2}{*}{5} & Fashion++ & 14 & 28 \\
 & Ours & \textbf{17} & \textbf{16} \\
\midrule
\multirow{2}{*}{Total} & Fashion++ & 44 & 143 \\
 & Ours & \textbf{121} & \textbf{51} \\
\bottomrule
\end{tabular}
}
\caption{The results of our user study. The numbers indicate the number of images rated as having increased or decreased fashionability. 50 pairs between input and Fashion++'s output and 50 pairs between input and our method's output are assigned to each participant. There are a total of 250 responses for each of these two kinds of comparison, respectively.} 

\label{tab:user_study}
\end{table}

\subsection{Negative Results}
\begin{figure}  \includegraphics[width=\linewidth]{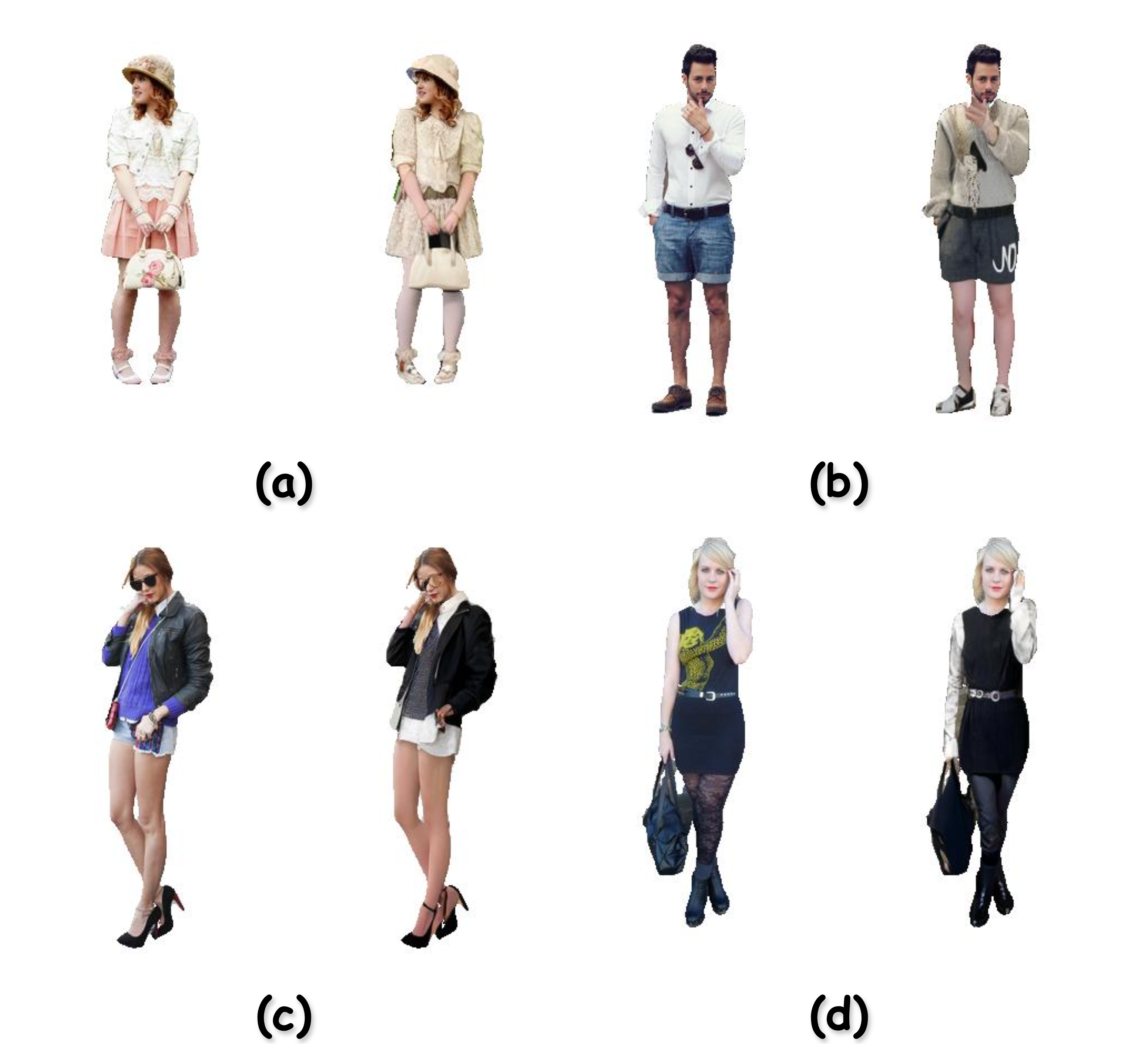}
   \caption{Examples of failure cases of our approach. For each pair of images, the left image represents the input image, and the right one shows the generated image of our approach. (a) and (b) highlight inconsistencies within the generated garments, where certain areas exhibit significantly different patterns or textures compared to the rest of the outfit. (c) and (d) demonstrate issues in generating body parts, such as missing hands or incorrect orientation of legs, revealing limitations in preserving accurate anatomical details.}
  \label{fig:badexample}
\end{figure}

As shown in Figure ~\ref{fig:badexample}, our approach occasionally fails to generate consistent and anatomically accurate results. In example (a), the woman is holding a bag in front of her dress, with the strap encircling part of the garment, but the enclosed area displays a markedly different texture and color. In example (b), the man’s glasses, hanging on his chest, are rendered as a vague patch rather than realistically integrated. These issues suggest that the model struggles to process accessory-garment interactions and small accessories due to a lack of specific annotations and limited training data for such cases. In example (c), the woman’s hand merges with her pants, creating an unnatural white patch, while in example (d), the crossed legs are distorted, resulting in an implausible pose. These errors highlight the model's challenges in maintaining realistic body structures for complex poses or overlapping body parts, likely due to ambiguous segmentation maps or insufficient pose diversity during training.

\section{Limitations and Discussions}

In this paper, we introduced a novel approach to enhance the fashionability of garment images by combining Classifier Guidance and ControlNet.
Our method leverages a custom-trained classifier and a specialized ControlNet pipeline to generate fashionably improved images, addressing the limitations of existing methods.
By training two distinct classifiers for fashionability prediction—one based on OpenSkill and another utilizing detailed expert ratings across multiple fashion aspects—we have established a robust framework for evaluating fashion images.
Extensive evaluations demonstrate that our method consistently surpasses the existing approach in generating visually appealing and fashion-forward images.

While our model significantly improves fashionability and generates high-quality images, several limitations remain. First, fashionability is inherently subjective, and despite using expert-annotated data, the generated outputs may not fully align with individual user preferences. Second, though effective, the proprietary dataset used for training is limited in size and diversity, which could impact the model's robustness in representing a wider range of styles and trends. Third, shading preservation on generated garments remains inconsistent, likely due to the model’s focus on enhancing fashionability rather than maintaining precise lighting consistency. Lastly, our method relies on segmentation maps that constrain garment dimensions to the original body proportions, limiting flexibility in modifying shapes and designs. Addressing these limitations in future work could further enhance the versatility and applicability of our approach in the fashion industry, contributing to its evolution in both virtual and real-world contexts.

\section*{Acknowledgments}

We thank Takuma Nakamura, Kazuya Morishita, and Sai HtaungKham for their support for data collection.

{\small
\bibliographystyle{ieee_fullname}
\bibliography{egbib}
}

\end{document}